\documentclass[conference]{IEEEtran}
\IEEEoverridecommandlockouts
\usepackage{cite}
\usepackage{amsmath,amssymb,amsfonts}
\usepackage{algorithmic}
\usepackage{graphicx}
\usepackage{textcomp}
\usepackage{xcolor}
\usepackage{cite}
\usepackage{subfigure}
\usepackage{bm}
\usepackage{amsmath}
\def\BibTeX{{\rm B\kern-.05em{\sc i\kern-.025em b}\kern-.08em
    T\kern-.1667em\lower.7ex\hbox{E}\kern-.125emX}}
\begin{document}

\title{A Multi-stream Convolutional Neural Network
	For Micro-expression Recognition Using Optical Flow and EVM\\}

\author{Jinming Liu, Ke Li, Baolin Song, Li Zhao\\
School of Information Science and Technology\\
Southeast University, Nanjing, P. R. China\\
Email: 213162988$@$seu.edu.cn}

\maketitle

\begin{abstract}
Micro-expression (ME) recognition plays a crucial role in a wide range of applications, particularly in public security and psychotherapy. Recently, traditional methods rely excessively on machine learning design and the recognition rate is not high enough for its practical application because of its short duration and low intensity. On the other hand, some methods based on deep learning also cannot get high accuracy due to problems such as the imbalance of databases. To address these problems, we design a multi-stream convolutional neural network (MSCNN) for ME recognition in this paper. Specifically, we employ EVM and optical flow to magnify and visualize subtle movement changes in MEs and extract the masks from the optical flow images. And then, we add the masks, optical flow images, and grayscale images into the MSCNN. After that, in order to overcome the imbalance of databases, we added a random over-sampler after the Dense Layer of the neural network. Finally, extensive experiments are conducted on two public ME databases: CASME II and SAMM. Compared with many recent state-of-the-art approaches, our method achieves more promising recognition results.
\end{abstract}


\section{INTRODUCTION}
Micro-expression refers to a rapid, unconscious, spontaneous facial movement that occurs when experiencing a strong emotion. It usually has a short maintenance time, between 0.2s to 0.04s. Unlike ordinary expressions, spontaneous expressions cannot be faked or suppressed but can reflect people's real emotions so that the reliability of micro-expression in emotion recognition task is very high. Therefore there are many applications of micro-expressions, such as lie detection, negotiating, marketing, distance learning, and so on. However, its duration is short, and change is subtle, usually appearing only either the upper or lower half of the face, but not the whole face\cite{lie}. Because of these problems, it is difficult to observe micro-expressions correctly.

In the past, methods based on Local Binary Pattern (LBP), optical flow, and machine learning were the main means of micro-expression recognition. They usually use LBP-based or optical flow methods to extract features and then classify them by Support Vector Machine (SVM). Most existing databases\cite{CASMEII}\cite{SAMM} use LBP with Three Orthogonal Planes (LBP-TOP)\cite{LBP-TOP} as their primary baseline feature extractor. Wang $et\, al.$\cite{LBP-SIP} proposed LBP-SIP that provided a lightweight representation based on three intersecting lines crossing over the center point of LBP-TOP. 
In addition to the LBP-TOP based approach, there are other approaches, such as, Xu $et\, al.$\cite{fdm} took the optical flow field as the basic feature to describe the motion pattern of micro-expressions and further extracted more simplified expression forms to propose facial dynamic map (FDM).

\begin{figure*}[tp]
	\centerline{\includegraphics[scale=0.23]{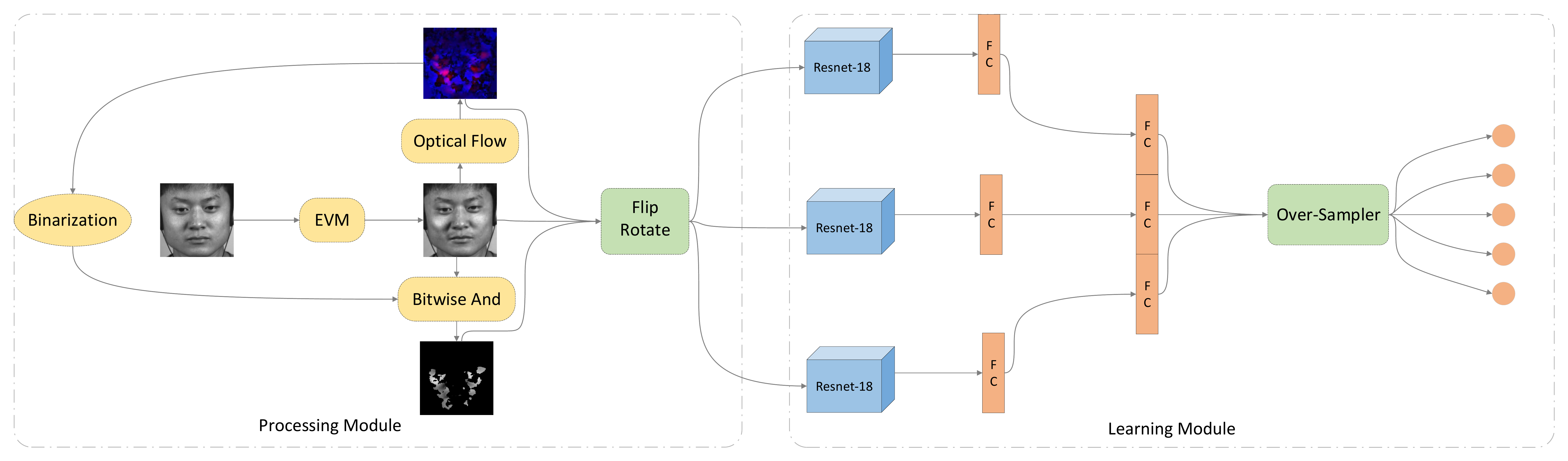}}
	\caption{Proposed framework}
	\label{framework}
\end{figure*}

Additionally, not just traditional machine learning, but deep learning can also be used for micro-expressions. In recent years, CNN has made great achievements in the fields of computer vision. Of course it could also be used for micro-expression recognition. Patel $et\, al.$\cite{CNN1} proposed a method based on temporal interpolation and CNN. Khor $et\, al.$\cite{khor2018enriched} proposed a three-stream neural network and an enrichment method to fuse various features. However, compared with traditional machine learning, deep learning is less used to recognize micro-expression. And there are some problems to be solved in these CNN models, such as the imbalance of databases, and the inputs need to be more effective features rather than just video sequences or optical flows (dynamic features) or grayscale images (static features).

\par 
To solve these problems, in this paper, we propose a method based on deep learning for micro-expression recognition. Firstly, we use EVM to magnify micro-expression. And then we use the optical flow calculation method\cite{Liu_PHD_2009} proposed by Bob $et \, al.$ to get the optical flow that contain information about the movements of facial muscles. After that, the optical flow images obtained are binarized with 50 as the threshold to obtain masks. After this step, we add the masks to the grayscale images to get the local areas of the face that have motion changes. At last, we input optical flow images (dynamic features), original gray images (static features) and the gray images which are covered by the masks (local features) into the multi-stream convolutional neural network proposed by ourselves. Different from the previous multi-stream neural network\cite{khor2018enriched}, we add an over-sampler after the network's final Dense Layer to deal with the problem of imbalanced databases.

\par 
Our main contributions are summarized as follows:

\begin{itemize}
	\item We combine EVM and optical flow method to process micro-expression. Comparing to using only one of the two, our method has higher accuracy.
	\item We propose a method to extract local features of motion and reduce data redundancy.
	\item We attempt to deal with the imbalance problem by adding an over-sampler inside the neural network
\end{itemize}

\par 
The rest of this paper is organized as follows. Section \ref{sec: III} presents our proposed method, and it is divided into two subsections to describe two modules: preprocessing and CNN. 
The preprocessing module includes steps such as extracting features and extending databases, while the other module determines some parameters of the deep network.
Then we discuss two related databases and the experimental results for algorithm evaluation in Section \ref{EXP}. Finally, Section \ref{sec: V} concludes this work.

\begin{table*}[htbp]
	\caption{INFORMATION ABOUT CASME-II AND SAMM}
	\begin{center}
		\setlength{\tabcolsep}{1mm}\begin{tabular}{|c|c|c|c|c|c|c|c|}
			\hline
			\textbf{Database}& Subjects & Samples & FPS& Elicitation & FACS coded & Frame annotations & Remarks\\
			\hline
			\textbf{CASME-II} & 26 & 247 & 200&Spontaneous & Yes & Onset, offset, apex & Disgust, Happiness, Repression, Surprise, and Others\\
			\hline
			\textbf{SAMM$^{\mathrm{*}}$} & 27 & 136 & 200 &Spontaneous & Yes & Onset, offset, apex & Anger, Contempt, Happiness, Surprise, and Others\\
			\hline
			\multicolumn{8}{l}{$^{\mathrm{*}}$We removed the three emotion classes of fear, sadness, and disgust in the original SAMM database}
		\end{tabular}
		
		\label{tab3}
	\end{center}
\end{table*}

\section{PROPOSED METHOD}
\label{sec: III}
In this section, we will introduce our proposed method of recognizing micro-expressions. Figure \ref{framework} summarizes the proposed framework with the preprocessing module and learning module.


\subsection{Pre-processing}
The micro-expression videos are first preprocessed using Eulerian Video Magnification (EVM), which could magnify the expression, makes it easier to recognize. However, the linear EVM method will amplify the noise while amplifying the changes in the movement so that we need  a reasonable amplification factor $\alpha$, and we will discuss this parameter in Section \ref{EXP}.

\par 
After using EVM to magnify expressions, we used the optical flow calculation method proposed by Liu $et\, al.$\cite{Liu_PHD_2009} to calculate optical flow.
They calculate dense optical flow, which computes the motion of each pixel, and the objective function shows in (\ref{equ:equ1}).
The image lattice is represented as $\mathbf{p}=(x,y,t)$ and flow is $\mathbf{w(\mathbf{p})}=(u(\mathbf{p}),v(\mathbf{p}),1)$, $\psi(\cdot)$ and $\phi(\cdot)$ are robust functions.

\begin{equation}
\begin{split}
E(du,dv)=&\int \psi\big(|I(\mathbf{p}+\mathbf{w}+d\mathbf{w})-I(\mathbf{p})|^2\big)\\
&+\alpha \phi\big(|\nabla(u+du)|^2+|\nabla(v+dv)|^2\big)d\mathbf{p}
\label{equ:equ1}
\end{split}
\end{equation}

\par
\begin{figure}[bp]
	\centering
	\subfigure[]{
		\begin{minipage}[t]{0.17\linewidth}
			\centering
			\includegraphics[scale=0.2]{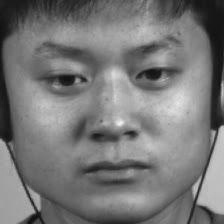}
			\label{onset}
		\end{minipage}%
	}%
	\subfigure[]{
		\begin{minipage}[t]{0.17\linewidth}
			\centering
			\includegraphics[scale=0.2]{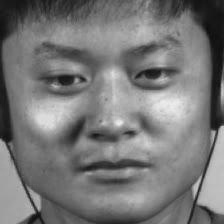}
			\label{apex}
		\end{minipage}%
	}%
	\subfigure[]{
		\begin{minipage}[t]{0.155\linewidth}
			\centering
			\includegraphics[scale=0.2]{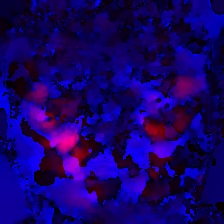}
			\label{img: optical}
		\end{minipage}
	}%
	\subfigure[]{
	\begin{minipage}[t]{0.155\linewidth}
		\centering
		\includegraphics[scale=0.2]{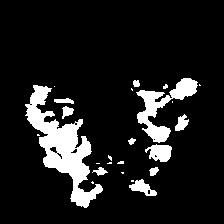}
		\label{mask}
	\end{minipage}
	}%
	\subfigure[]{
	\begin{minipage}[t]{0.17\linewidth}
		\centering
		\includegraphics[scale=0.2]{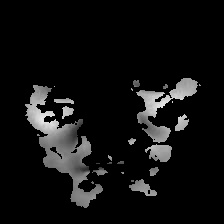}
		\label{masked}		
	\end{minipage}
    }%
	\centering

	\caption{Images of EP02-01f from CASME-II. (a) Micro-expression onset frame; (b) Apex frame; (c) The optical flow which is calculated from the images of frames onset and apex; (d) Binary optical flow image; (e) Grayscale image covered by a mask generated by binary optical flow image}
\end{figure}

 Then we will get the optical flow of Figure \ref{apex} and Figure \ref{onset}, the visual image is shown in Figure \ref{img: optical}.
In our proposed method, we use the onset frame, apex frame, and apex frame to generate optical flow.

\par 
Xia $et\, al.$ \cite{xia2019spatiotemporal} proposed a method to extract mask based on the heatmap. The mask produced by this method included both the upper and lower parts of the face, whereas ME usually appeared on only half the face\cite{lie}. So we binarized the optical flow image with a threshold value of 50 to get a mask, this method will only extract the moving part, as the Figure \ref{mask} shows. This mask is used to be added to the apex frame of a video sequence, and it can extract the effective areas of motion in the image and remove some redundancy. The result is shown in Figure \ref{masked}.

\par 
Due to the small number of ME database samples will lead to bad robustness, to improve the robustness of the model and the reliability of the results, in the last step of preprocessing, we expand the training samples, using horizontal inversion and clockwise/counterclockwise rotation of 5°/10° to expand the number of samples by 10 times.

\subsection{Convolutional Neural Network (CNN)}
In this paper, we use Resnet-18 \cite{he2016deep} to learn training data. It takes images size of $224\times224$ and batch normalization is carried out before each convolutional layer for faster training convergence. All convolutional and  fully connected layers adopt Rectified Linear Unit (ReLU) as the activation function. 
Micro-expressions often overfit because of its small databases, so here we use Early Stopping to mitigate these overfitting problems. The classifier we used is Softmax, then we used a mini-batch size of 64, and the maximum epoch is 200. The learning rate is $10^{-3}$, and momentum is 0.9. The weight decay is $5\times10^{-4}$, and optimizer is SGD. The network's three inputs are optical flow images ($O \in \mathbb{R}^3$), gray-scale raw images ($G \in \mathbb{R}^2$) and gray-scale images that are covered by masks ($M \in \mathbb{R}^2$).
%
%
%
%

\begin{table}[t]
	\caption{PERFORMANCE OF PROPOSED METHODS VS. OTHER METHODS FOR
		MICRO-EXPRESSION RECOGNITION ON CASME-II}
	\begin{center}
		\setlength{\tabcolsep}{1mm}\begin{tabular}{|c|c|c|}
			\hline
			\textbf{Methods}& \textbf{F1-Score} & \textbf{Accuracy} \\
			\hline
			ELRCN-TE\cite{khor2018enriched} & 0.5000  & 0.5244\\
			\hline
			LBP-SIP\cite{LBP-SIP} & 0.4480 & 0.4656\\
			\hline
			CNN + SFS\cite{CNN1} & N/A & 0.4730\\
			\hline
			Adaptive MM + LBP-TOP\cite{park2015subtle} & N/A  & 0.5191 \\
			\hline
			FDM\cite{fdm} & 0.4053 & 0.4593\\
			\hline
			\textbf{	$\mathbf{Ours}$ (EVM + Optical) }&  0.4705 & $0.5077$\\
			\hline
			\textbf{	$\mathbf{Ours}$  (MSCNN) }& $\bm{0.5325}$  & \textbf{0.5650}\\
			\hline
			
		\end{tabular}
		
		\label{tab2}
	\end{center}
\end{table}

\section{EXPERIMENT}
\label{EXP}
\subsection{Databases and experiment setting}
At present, the commonly used databases of ME are CASME-II\cite{CASMEII} from Chinese academy of sciences and SAMM database\cite{SAMM}. 
\par
CASME-II is a comprehensive spontaneous ME database containing 247 video samples, elicited from 26 Asian participants, and it is obtained using a high-speed camera of 200 frames per second. In order to ensure the reliability of the data set, CASME-II induced the subjects' emotions by watching videos, and if the subjects successfully suppressed their emotions without being noticed by the record, they could receive a cash reward.

\par

The SAMM database has 159 micro-expressions from 29 subjects. It has the same frame rate as CASME-II, and its induction mechanism is also similar to that of CASME-II.

\par 

CASME-II and SAMM have coded both emotions and FACS for each expression. CASME-II is divided into five emotion classes while SAMM is divided into eight emotion classes, but in order to be consistent with CASME-II, we only select the five classes with the largest number of samples for the experiment. Information about these two databases can be seen in Table\ref{tab3}.

\begin{figure*}[t]
	
	\centering
	\subfigure[]{
		\begin{minipage}[t]{0.3\linewidth}
			\centering
			\includegraphics[scale=0.22]{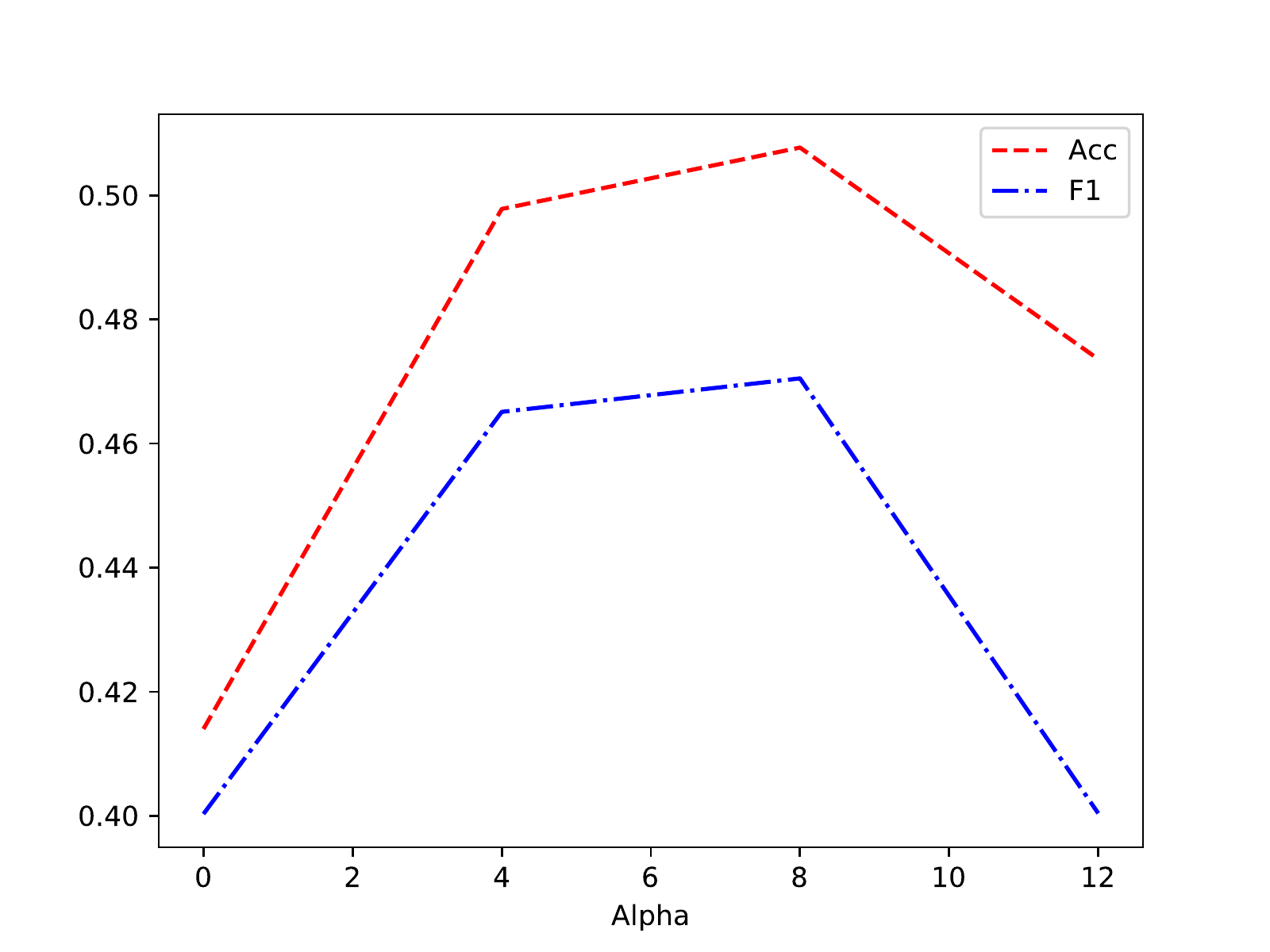}
			\label{fig:alpha}
		\end{minipage}%
	}%
	\subfigure[]{
		\begin{minipage}[t]{0.3\linewidth}
			\centering
			\includegraphics[scale=0.22]{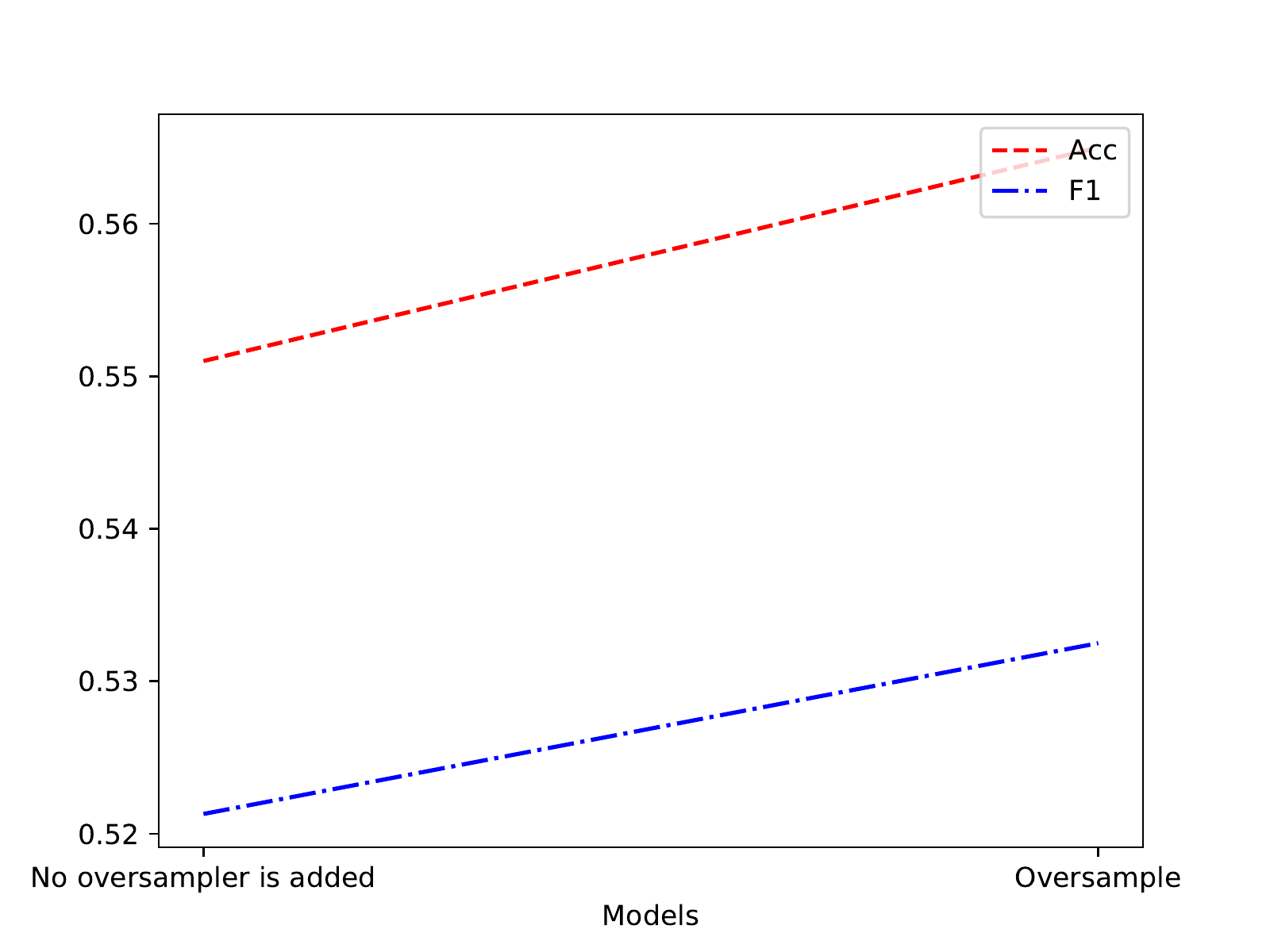}
			\label{fig:up}
		\end{minipage}%
	}%
	\subfigure[]{
		\begin{minipage}[t]{0.3\linewidth}
			\centering
			\includegraphics[scale=0.22]{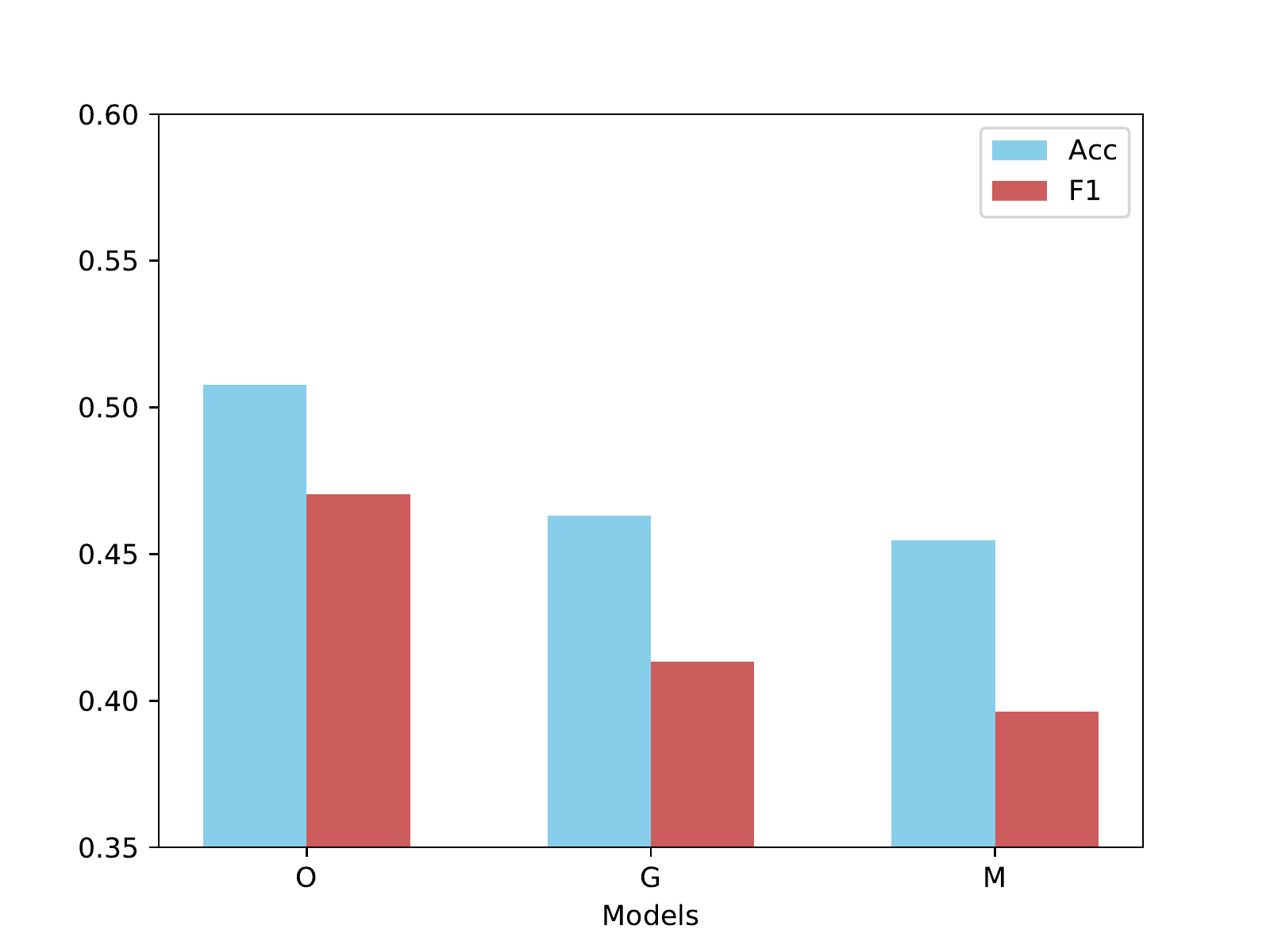}
			\label{fig:single}
		\end{minipage}%
	}%
	
	\subfigure[]{
		\begin{minipage}[t]{0.3\linewidth}
			\centering
			\includegraphics[scale=0.22]{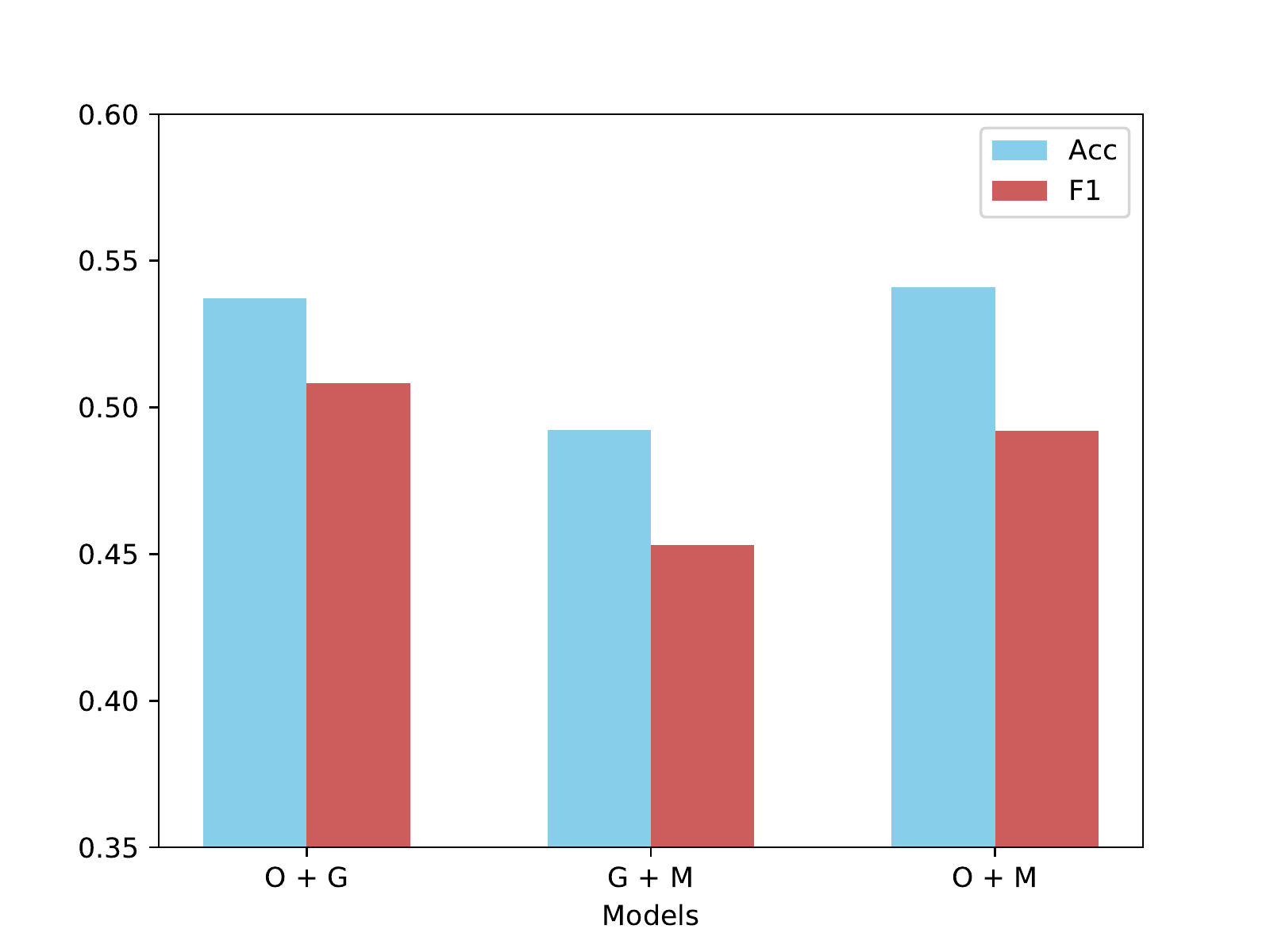}
			\label{fig: double}
		\end{minipage}%
	}
	\subfigure[]{
		\begin{minipage}[t]{0.3\linewidth}
			\centering
			\includegraphics[scale=0.22]{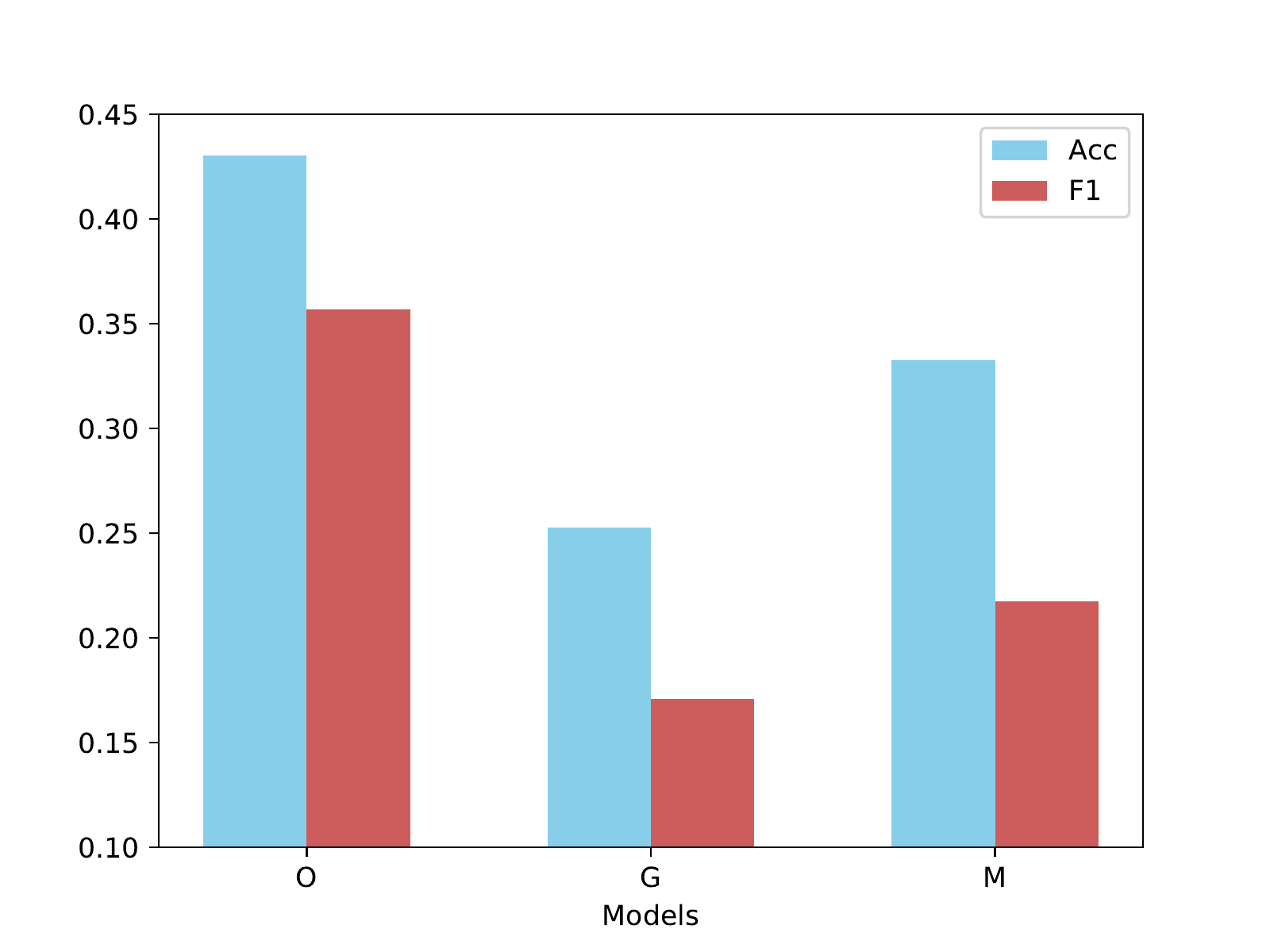}
			\label{fig: ssingle}
		\end{minipage}%
	}%
	\subfigure[]{
		\begin{minipage}[t]{0.3\linewidth}
			\centering
			\includegraphics[scale=0.22]{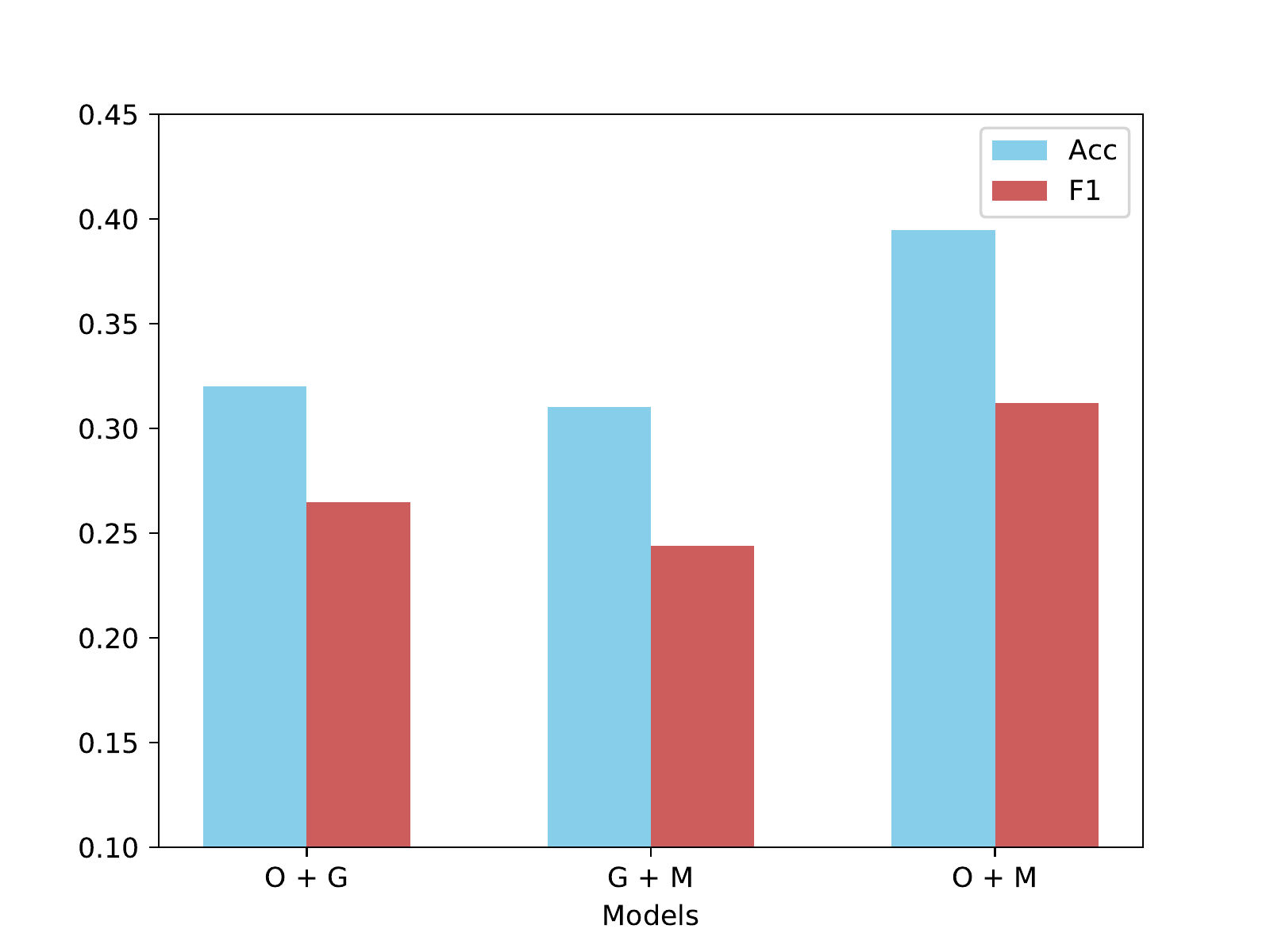}
			\label{fig: sdouble}
		\end{minipage}%
	}%
	
	\caption{The accuracy and F1-Score of CASME-II and SAMM. (a) The accuracy and F1-Score of different $\alpha$ parameters; (b) No over-sampler is added and Containing an Over-sampler; (c) Single-stream based on CASME-II; (d) Double-stream based on CASME-II; (e) Single-stream based on SAMM; (f) Double-stream based on SAMM}
\end{figure*}

\par

\par 
 In order to avoid too much distortion, we need to set a reasonable amplification factor $\alpha$. To find the best $\alpha$ for EVM, we tried several different sets of values and calculated the optical flow of the CASME-II database under this value. The final results are shown in Figure \ref{fig:alpha}. As we can see, the best result is obtained when the parameter is 8. 
\subsection{Comparison Results on CASME-II and SAMM}
\begin{figure}[tp]
	\centerline{\includegraphics[scale=0.48]{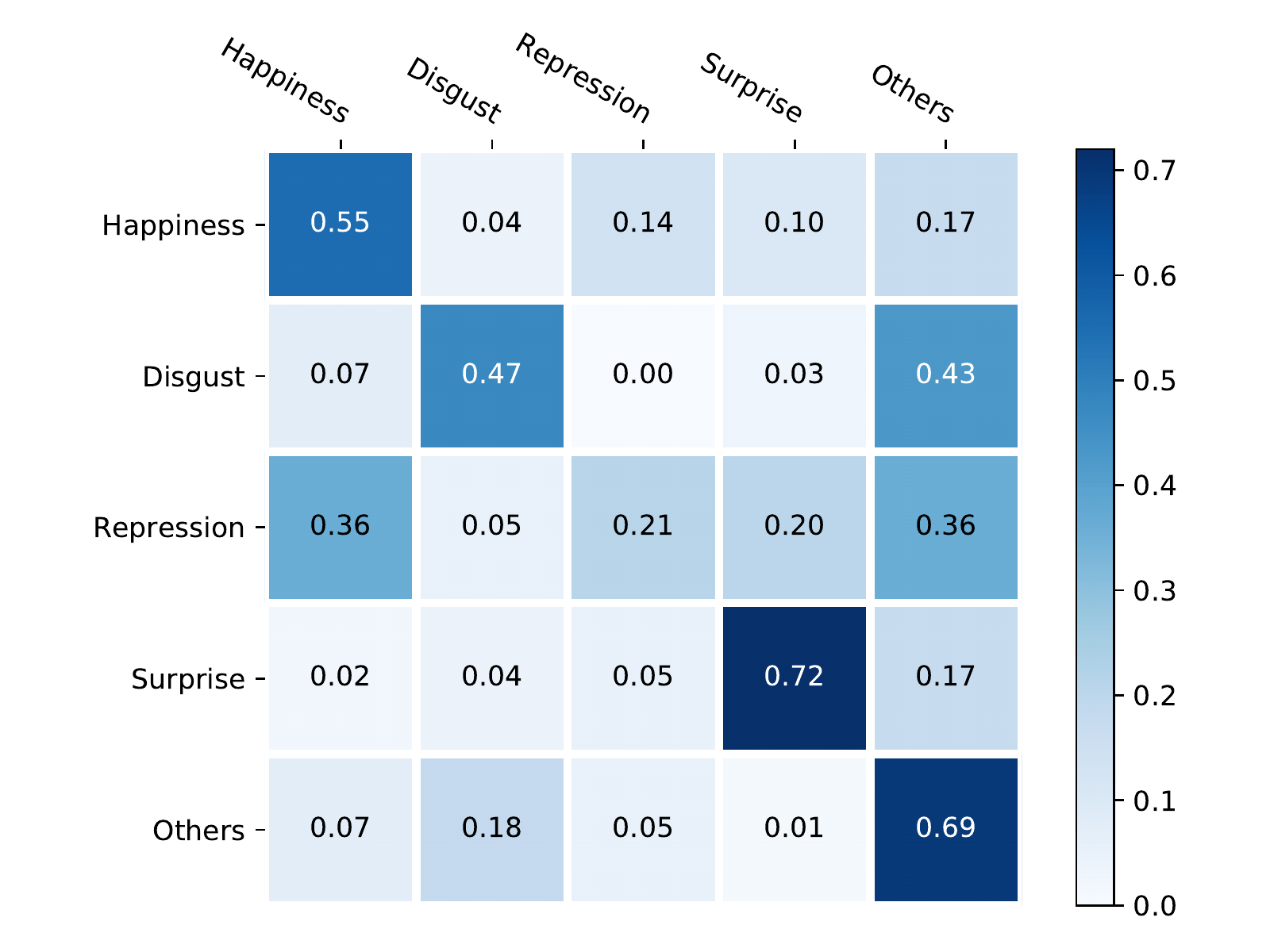}}
	\caption{Confusion Matrix of our proposed method on LOSO protocol}
	\label{fig:ma}
\end{figure}

We evaluated our approach on two spontaneous facial ME databases: CASME-II and SAMM using the Leave-One-Subject-Out (LOSO) cross validation as this protocol prevents subject bias during learning. Table \ref{tab2}, \ref{tab4} compares the performance of our proposed approach  with the baseline LBP-TOP approach and some recent and related works. The confusion matrix is shown in Figure \ref{fig:ma}. It can be seen that our method has the highest accuracy and F1-Score among these methods, which is not only superior to some of the previously mentioned deep learning methods but also higher than some traditional machine learning methods. And for the SAMM database, we got a result that is lower than CASME-II, and the result of the multi-stream neural network is not as good as that of single-stream optical flow, and this may be because the SAMM database is relatively small, resulting in poor results of the other two streams, which will also lower the performance of optical flow after feature fusion. However, since this over-sampler is a random over-sampler, it may cause overfitting, so it does not perform well in the SAMM database.

\begin{table}[t]
	\caption{PERFORMANCE OF PROPOSED METHODS VS. OTHER METHODS FOR
		MICRO-EXPRESSION RECOGNITION ON SAMM}
	\begin{center}
		\setlength{\tabcolsep}{1mm}\begin{tabular}{|c|c|c|}
			\hline
	      \textbf{Methods}& \textbf{F1-Score} & \textbf{Accuracy} \\

			\hline
			HOOF\cite{hoof} & 0.33 & 0.4217\\
			\hline
			HOG3D\cite{hog3d} & 0.22 & 0.3416\\
			\hline
			LBP-TOP\cite{LBP-TOP} & 0.1768& 0.3556\\
			\hline
				      	\textbf{	$\mathbf{Ours}$ (MSCNN) }&  0.3173& $0..4071 $\\		
				      	\hline
			\textbf{	$\mathbf{Ours}$  (EVM + Optical) }& \textbf{0.3569 } & $\bm{0.4304}$\\
			\hline
			
		\end{tabular}
		
		\label{tab4}
	\end{center}
\end{table}

\par 

\par 
To overcome the imbalance of databases, we added the over-sampler, to verify the validity of this step, we compared the results obtained by no sampler with those obtained by a three-flow neural network adding a sampler based on CASME-II, as is shown in the Figure \ref{fig:up}.  However, since this over-sampler is a random over-sampler, it may cause overfitting, so it does not perform well in the SAMM database.

\par 
For further analysis, we also analyzed the different single-stream and double-stream inputs. These are carried out using both the CASME-II database and SAMM database, and are shown in Figure \ref{fig:single}, \ref{fig: double}, \ref{fig: ssingle}, \ref{fig: sdouble}.

\section{CONCLUSION}
\label{sec: V}

In this paper, we propose a MSCNN for ME recognition using optical flow and EVM. Firstly, the architecture of MSCNN is suitable for dealing with MEs recognition problem, because it can fuse different features, get more effective information. Secondly, it is clear that we use EVM, optical flow and a mask to extract feature, when $\alpha$ is 8, which benefit distinguishing MEs as well. Finally, we add an oversampler to reduce the effect of imbalanced database.  Through experiments, we find that MSCNN has better performance than other methods mentioned. In the future, we will try to use some other over-samplers, such as SMOTE, ADASYN over-sampler and so on. In addition, we will focus on a network structure more suitable for ME recognition than the residual network.


\end{document}